%% file: root.tex
\documentclass[letterpaper, 10 pt, conference]{ieeeconf}  

\IEEEoverridecommandlockouts                              

\usepackage{graphicx} 
\usepackage{subcaption}
\usepackage{amsmath} 
\usepackage{amssymb}  
\usepackage{cite}
\usepackage{cleveref}

\crefname{figure}{Fig.}{Figs.}
\Crefname{figure}{Fig.}{Figs.}

\crefname{table}{Table}{Tables}
\Crefname{table}{Table}{Tables}

\crefname{section}{Section}{Sections}
\Crefname{section}{Section}{Sections}

\title{\LARGE \bf
Autonomous Legged Mobile Manipulation for Lunar Surface Operations via Constrained Reinforcement Learning*
}

\author{Alvaro Belmonte-Baeza$^{1}$, Miguel Cazorla$^{1}$, Gabriel J. García$^{2}$, Carlos J. Pérez-Del-Pulgar$^{3}$, and Jorge Pomares$^{2}$
\thanks{*This research was suported by the project PID2024-160373OB-C22 funded by MICIU /AEI /10.13039/501100011033 / FEDER, UE and by the grant FPU21/02586.}
\thanks{$^{1}$Alvaro Belmonte-Baeza and Miguel Cazorla are with the Department of Computer Sciences and Artificial Intelligence, University of Alicante, Spain
        {Corresponding author: \tt\small alvaro.belmonte@ua.es}}%
\thanks{$^{2}$Gabriel J. Garcia and Jorge Pomares are with the Department of Physics, Systems Engineering, and Signal Theory, University of Alicante, Spain.
        {}}%
\thanks{$^{3}$Carlos J. P\'erez-Del-Pulgar is with the Department of Systems and Automatics Engineering, University of M\'alaga, Spain
        {}}%
}

\begin{document}

\maketitle

\begin{abstract}

Robotics plays a pivotal role in planetary science and exploration, where autonomous and reliable systems are crucial due to the risks and challenges inherent to space environments. The establishment of permanent lunar bases demands robotic platforms capable of navigating and manipulating in the harsh lunar terrain. While wheeled rovers have been the mainstay for planetary exploration, their limitations in unstructured and steep terrains motivate the adoption of legged robots, which offer superior mobility and adaptability. This paper introduces a constrained reinforcement learning framework designed for autonomous quadrupedal mobile manipulators operating in lunar environments. The proposed framework integrates whole-body locomotion and manipulation capabilities while explicitly addressing critical safety constraints, including collision avoidance, dynamic stability, and power efficiency, in order to ensure robust performance under lunar-specific conditions, such as reduced gravity and irregular terrain. Experimental results demonstrate the framework's effectiveness in achieving precise 6D task-space end-effector pose tracking, achieving an average positional accuracy of 4 cm and orientation accuracy of 8.1 degrees. The system consistently respects both soft and hard constraints, exhibiting adaptive behaviors optimized for lunar gravity conditions. This work effectively bridges adaptive learning with essential mission-critical safety requirements, paving the way for advanced autonomous robotic explorers for future lunar missions.

\end{abstract}

\input{sections/introduction}
\input{sections/background}
\input{sections/methodology}
\input{sections/results}

\input{sections/conclusions}




\section*{ACKNOWLEDGMENT}

Alvaro Belmonte-Baeza thanks Elliot Chane-Sane for his thorough explanations and help during our custom CaT implementation for IsaacLab.


\bibliographystyle{IEEEtran}
\bibliography{references}

\end{document}

%% file: sections/introduction.tex
\section{INTRODUCTION}\label{sec:intro}

Robotic systems are indispensable for modern planetary exploration, where autonomous operation, precision, and reliability are essential due to the extreme risks and communication delays associated with space missions. The forthcoming establishment of permanent lunar infrastructures and the exploitation of in-situ resources require robotic platforms capable of performing complex manipulation and mobility tasks in environments that are not only remote but also extremely unforgiving. Missions such as NASA’s Artemis \cite{nasa_artemis_2025} and China’s Chang’e \cite{JIA2018207} programs highlight this growing reliance on robotic systems for long-duration lunar surface operations.

\begin{figure}[t]
    \centering
    \includegraphics[width=\columnwidth]{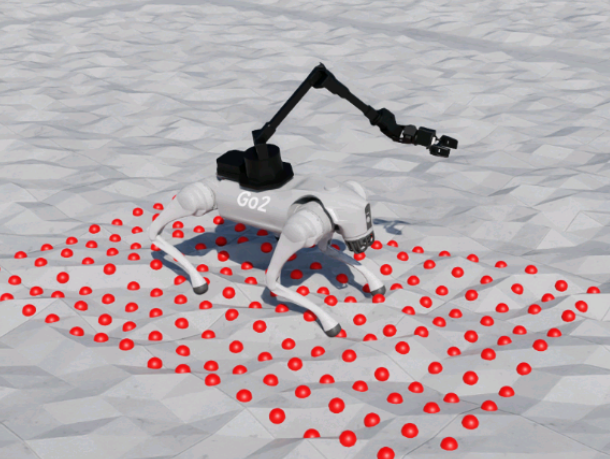}
    \caption{Legged Mobile Manipulator in Lunar environment used in our work}
    \label{fig:intro}
\end{figure}

To date, planetary exploration has relied primarily on wheeled rovers, which have demonstrated remarkable robustness and simplicity. However, their performance degrades severely on steep slopes, uneven surfaces, and soft regolith—conditions that dominate the lunar landscape. Consequently, the planetary robotics community is shifting its attention toward legged robots, which offer superior adaptability and terrain accessibility. Several research prototypes, including SpaceClimber \cite{5756948}, ATHLETE \cite{Wilcox2007}, and SpaceBok \cite{8794136, Kolvenbach2022}, have explored various aspects of legged mobility for space exploration, from statically stable climbing to highly dynamic gaits optimized for reduced-gravity locomotion. These developments illustrate that legged systems can extend the reach of planetary missions into areas previously inaccessible to wheeled rovers.

Beyond locomotion, future lunar operations will demand robots capable not only of traversing rugged terrain but also of performing manipulation tasks such as sample retrieval, habitat construction, and equipment maintenance. Quadrupedal mobile manipulators represent an especially promising solution, as they combine the mobility of legged locomotion with the dexterity of articulated arms (see \cref{fig:intro}). Coordinating these two subsystems, however, introduces a high-dimensional control problem requiring real-time whole-body reasoning about balance, forces, and safety constraints.

Reinforcement learning (RL) has recently emerged as a powerful paradigm for enabling robust and adaptive control behaviors in high-dimensional robotic systems. RL-based controllers have achieved state-of-the-art performance in quadrupedal and humanoid locomotion \cite{HoellerAnymalParkour, MikiWildAnymal, RLHumanoid}, dexterous manipulation \cite{openai2019solvingrubikscuberobot}, and integrated navigation \cite{rudin2022advancedskillslearninglocomotion, shah2022viking}. More recently, several works have addressed whole-body loco-manipulation, where locomotion and manipulation are learned jointly to accomplish coordinated tasks. For example, joint control of base and arm is shown in \cite{fu2022deep, ha2024umilegs}, although manipulation and locomotion are handled separately in the first case, and manipulation is carried by a separate computationally-intensive diffusion policy in the second case. Other approaches do integrate end-effector tracking and balance control to achieve adaptive behaviors, both in flat and irregular terrain \cite{portela2025wholebodyendeffectorposetracking, He2024, Li_2024_11}. However, these frameworks typically focus on terrestrial environments and lack mechanisms to guarantee formal safety properties, an essential requirement for autonomous operation in planetary conditions where failures are irrecoverable.

These advances in learning-based whole-body control are increasingly being recognized as essential for enabling adaptive and intelligent behavior in planetary robotic systems due to their ability to learn from experience and adapt to uncertain environments. Furthermore, the neural network architectures used by these policies are often simple Multi-Layer Perceptrons (MLP) that can be deployed with more modest computational resources. However, the lack of formal safety guarantees that learning-based controllers have is still holding back the space robotics community from applying these techniques to the planetary domain. 

Operating on the lunar surface presents unique challenges that go beyond those encountered on Earth. The Moon’s reduced gravity (approximately one-sixth of Earth’s), combined with highly irregular topography, loose regolith, and the absence of atmosphere, creates conditions where small control instabilities can quickly lead to catastrophic failure. Moreover, limited communication bandwidth and time delays preclude direct human supervision. Consequently, safety-aware autonomy is critical: lunar robots must be capable of self-preserving operation, maintaining stability and efficiency even when facing disturbances or uncertainties in terrain properties.

In this context, Constrained Reinforcement Learning (CRL) offers a principled framework for combining the adaptive power of RL with explicit safety guarantees. Unlike conventional RL, which optimizes a reward function alone, CRL incorporates task-specific constraints that the learned policy must satisfy during both training and execution \cite{achiam2017constrained}. This paradigm has shown promise for improving safety and robustness in terrestrial legged locomotion \cite{LeeCRL, KimNotOnlyRewards, chane2024cat, chanesane2024solo}, and emerging loco-manipulation tasks \cite{dadiotis2025dynamicobjectgoalpushing, MaBadminton}. Yet, its application to planetary robotics, and in particular to legged mobile manipulators operating under lunar conditions, has not been explored until now.

This work introduces a Constrained Reinforcement Learning framework for autonomous legged mobile manipulators designed for lunar surface operations. Our framework enables integrated locomotion and manipulation control under lunar-specific environmental conditions, explicitly embedding safety-critical constraints such as collision avoidance, stability maintenance, and power efficiency into the learning process. By doing so, the method combines adaptive learning with formal safety considerations, two aspects that are rarely unified in planetary robotics.

The main contributions of our work include:
\begin{itemize}

\item \textbf{Whole-Body Coordination}: We propose an integrated control architecture that jointly optimizes locomotion and manipulation behaviors to achieve dynamic stability and coordinated motion, leveraging all system degrees of freedom in task execution.
\item \textbf{6D Task-space end-effector tracking}: Our CRL formulation enables precise 6D end-effector pose tracking in task space, achieving an average positional accuracy of 4 cm and orientation accuracy of 8.1°, comparable to the best terrestrial benchmarks.
\item \textbf{Safe Reinforcement Learning Architecture}: The method enforces explicit soft and hard constraints encompassing collision avoidance, torque and velocity limits, power efficiency, and body orientation safety—critical for autonomous operation in low-gravity environments.
\item \textbf{Environmental Adaptation}: The framework learns to operate over rough, uneven, and low-gravity terrain through domain randomization and stochastic constraint enforcement, demonstrating emergent behaviors such as energy-efficient gaits under lunar gravity.

\end{itemize}

Through these contributions, our approach bridges the gap between adaptive learning-based control and the reliability required for real-world lunar robotic missions. By explicitly incorporating safety constraints into the learning process, this work represents a significant step toward autonomous quadrupedal mobile manipulators capable of performing complex scientific and operational tasks in extraterrestrial environments.

%% file: sections/background.tex
\section{BACKGROUND}\label{sec:background}

\subsection{Constrained Reinforcement Learning Problem}\label{sec:crl}

In a Reinforcement learning framework, we generally model a sequential decision-making problem (such as robot control) as a Markov Decision Process (MDP). An MDP is defined by a tuple $(\mathcal{S}, \mathcal{A}, \mathcal{R}, \mathcal{P})$, where $\mathcal{S}$ is the state space, $\mathcal{A}$ is the action space, $\mathcal{R}: S \times A \times S \rightarrow \mathbb{R}$ corresponds to the reward function that maps a state transition via an action to a scalar reward, and $\mathcal{P}: S \times A \times S \rightarrow [0, 1]$ gives the transition probability from one state to another when an action $A$ is taken by the agent. The goal of the RL problem is to find a policy $\pi: S \rightarrow A$ that maximizes the expected cumulative reward.
\begin{equation}\label{eq:rl-objective}
    J(\pi) = \mathbb{E}\left[\Sigma_{t=0}^{\infty} \gamma^tr(s_t, a_t, s_{t+1})\right],
\end{equation}

where $\gamma \in [0,1)$ is known as the discount factor, which is used to ponderate the relative importance between short and long term rewards.

To address constrained sequential decision-making problems, the framework above can be extended into a Constrained MDP (CMDP). To do so, we introduce a set of constraints $\mathcal{C}$ represented as cost functions and associated limits $\mathcal{L}$. Thus, each of the defined constraints $c_i \in \mathcal{C}$ translates a state transition to the cost of that transition, and yields a cost function
\begin{equation}\label{eq:crl-cost}
    J_{c_i}(\pi) = \mathbb{E}\left[\Sigma_{t=0}^{\infty}\gamma^t c_i(s_t, a_t, s_{t+1}) \right].
\end{equation}

With these considerations, a constrained learning problem seeks to find a policy that maximizes the objective function described in \cref{eq:rl-objective} while maintaining the discounted sum of future costs $c_i$ within their defined limits $l_i$:

\begin{equation}\label{eq:crl-problem}
\begin{aligned}
\pi^* &= \arg\max_{\pi} J(\pi) \\
\text{s.t. } J_{c_i}(\pi) \leq & l_i \quad \forall i \in \{1, \dots, L\},
\end{aligned}
\end{equation}

As can be inferred by the formulation above, the constrained RL setting introduces a set of $\mathcal{C}$ cost functions, each of which needs to be addressed with separate critic networks, thus requiring specialized algorithm implementation and preventing the use of well-known off-the-shelf RL algorithm implementations.

\subsection{Constraints as Terminations}\label{sec:CaT}

The Constraints as Terminations (CaT) framework \cite{chane2024cat} aims to reduce the implementation complexity problem by reformulating the constrained RL objective, in order to prioritize simplicity and ease of use. To do so, stochastic terminations are introduced during policy learning to enforce the desired constraints, so that any violation of such constraints implies a probability of terminating the future rewards obtained from that timestep onwards. 

Looking back at \cref{eq:crl-problem}, the inequality introduces the concept of \textit{budget} or \textit{allowance} for constraints. This can be reformulated as maximizing rewards while avoiding violating a constraint such that $\mathcal{P}(s,a)\cdot[c_i(s,a) > 0] \leq \tilde{l_i}  \forall i \in L$. This simplification only holds for special cases of the general constrained RL framework, but it is sufficient for most applications of RL for robot control. 

Thus, the CaT framework reformulates the learning objective in \cref{eq:rl-objective} to:
\begin{equation}\label{eq:CaT}
    \max_{\pi} \mathbb{E}_{\tau \sim \pi} \left[\Sigma_{t=0}^{\infty} \left(\Pi_{t'=0}^t \gamma^{t'} (1 - \delta(s_{t'}, a_{t'}))\right) r(s_t, a_t)\right],
\end{equation}

where $\delta_t \in [0,1]$ is a random variable representing if the episode is terminated at timestep $t$, and consequently all future rewards are not received by the agent. This $\delta_t$ variable is of course a function of the constraints $c_i$, with its value depending on the violations of such constraints. Furthermore, the CaT formulation proposes that $\delta_t$ takes values within the [0,1] interval, resulting in a stochastic termination that probabilistically terminates future rewards based on constraint violations at the current timestep.

Following this reasoning, at each timestep $\delta$ is computed as per the following expression:
\begin{equation}\label{eq:cat-delta}
    \delta = \max_{i \in I} p_i^{max} \cdot clip\left(\frac{c_i^{+}}{c_i^{max}}, 0, 1\right),
\end{equation}

where $c_i^{+} = max(0, c_i(s,a))$ is the violation of $c_i$, and $c_i^{max}$ is a moving average of the maximum constraint violation that updates empirically during training over each batch of collected data. The clipping operation bounds the term $\frac{c_i^{+}}{c_i^{max}}$ within the $[0,1]$ interval.

With these modifications, the CaT framework allows for easy integration of state-of-the-art RL algorithms like PPO with the constrained RL framework. It is only necessary to weight the obtained rewards by a factor of $(1-\delta)$, and update the episode termination flags by $\delta$.

%% file: sections/methodology.tex
\section{METHOD}\label{sec:method}

We propose a Constrained Reinforcement Learning formulation of the legged locomanipulation problem to achieve safe and robust whole-body control of a robotic system consisting of a quadrupedal robot with a manipulator mounted on top of its body. The constraints imposed to the system condition the learning process so that violating any constraint results in the agent not receiving part of the reward collected during the time that the constraints are not being respected. This naturally makes the agent lean towards satisfying the constraints, in order to maximize the received reward.

\begin{figure*}[ht]
    \centering
    \includegraphics[width=\textwidth]{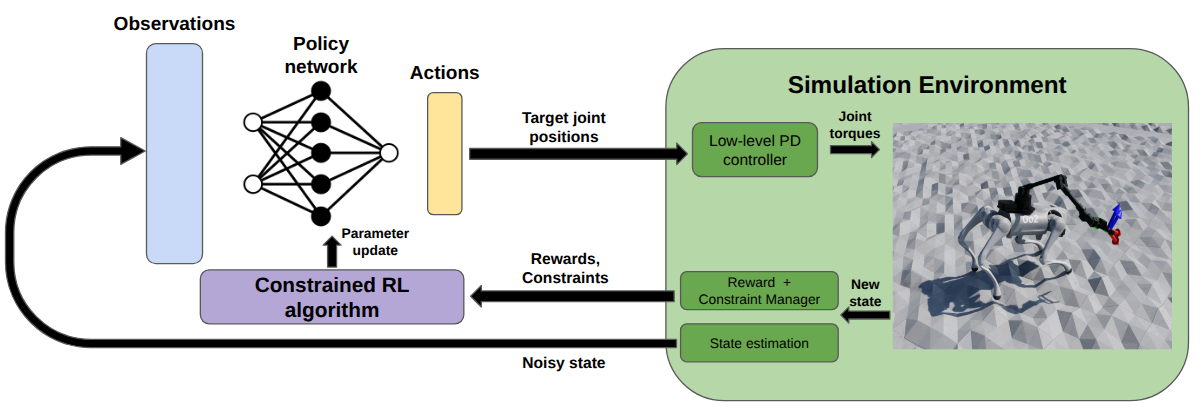}
    \caption{Overview of the proposed methodology.}
    \label{fig:overview}
\end{figure*}

In this section, we describe our framework for tracking a desired task-space 6D end-effector position via whole-body control of a legged locomanipulator system in Lunar environments. We first present the different parts of the Markov Decision Process to formally describe our sequential decision-making problem, with crucial details in the reward function formulation for a precise and efficient tracking of desired task-space pose. Subsequently, we will define the constraints imposed to our agent, in order to ensure that a set of safety guarantees are considered by the policy when learning the task at hand. An overview of the proposed methodology is depicted in \cref{fig:overview}, with each of the different modules being described in the following subsections.

\subsection{Whole-body Legged Locomanipulation Problem Formulation}

We aim to train a policy that is able to safely track a 6D end-effector pose in task space. Most previous works define the whole-body tracking problem with respect to the body frame of the robot, since this simplifies policy learning by working only with local information. However, if the body of the robot gets displaced for some reason, the end effector position will also change since the relative position between body and EE hasn't changed. This behavior is not desirable in the application scenarios we propose in \cref{sec:intro}, where we want to precisely manipulate or inspect certain areas given in task-space frame.

Thus, we must define a MDP that correctly formulates the desired task, namely tracking a desired EE pose with whole-body coordination. In the following paragraph, we detail our specific formulation to achieve the desired behavior.

\subsubsection{End-effector Pose Command in Task Space}\label{sec:command}

We define the command as a desired end-effector pose, $p_{ee} \in \mathbb{SE}(3)$ in task space. The command is sampled during learning as a random 6D pose within a vicinity of the robot, such that it might need to traverse some terrain to reach it, but won't need to move very far away. We argue that this definition suffices for our problem, since the policy learned here is not aimed for long-distance locomotion. Instead, we would deploy our policy when the robot is in a local area of interest that we want to inspect or retrieve some samples from.

\subsubsection{Action and Observation Spaces}\label{sec:spaces}

We define the action space as the desired joint positions for both the arm and the quadrupedal robot legs. In our case, the employed robotic arm has 6 DoF, while the quadrupedal robot has 4 legs with 3 joints each, resulting in $a_t \in \mathbb{R}^{18}$. The target positions outputted by the policy are then processed before being sent to two separate low-level PD controllers for the arm and the legs. The raw policy action is multiplied by a scaling factor, $\mathbf{\sigma}$, and added to the default joint position, such that $\mathbf{q}_{,t}^{*} = \mathbf{a}_t \cdot \mathbf{\sigma} + \mathbf{q_{default}}$. 

The observation space needs to include all relevant information for the locomanipulator to correctly perform the desired task-space tracking task. This includes both proprioceptive information to account for the robot state, as well as exteroceptive perception to adapt the motion to the rough lunar terrain that the robot needs to traverse in order to reach its desired EE pose. 

Thus, our observation vector consists of the robot body orientation defined by the projected gravity vector, as well as the body's linear and angular velocities. We also include joint positions and velocities for all the systems' joints to have complete joint-space state information, as well as the action applied by the policy in the previous step. We also add a four-dimensional boolean vector representing the contact state of each of the quadruped feet to assess the robot's stability and interaction with the environment. For exteroceptive perception, we add a height scan performed by the LIDAR mounted on the robot's chin that provides local terrain elevation information. 

Lastly, we feed the policy with the current desired EE pose in task space. Crucially, although the target is generated in task frame as described in \cref{sec:command}, we transform the desired EE pose to body frame before appending it to the observation vector. This is done to preserve the benefits that local information has for policy learning, while maintaining the task-space definition of the pose that a mission planner would provide.

For reference and notation clarity, \cref{tab:observations} shows the observations considered for our policy.

\begin{table}[t!]
\caption{Policy Observations}
\label{tab:observations}
\begin{center}
\begin{tabular}{c|c}
\hline
\bfseries \textbf{Observation name} & \bfseries \textbf{Expression} \\
\hline
Projected gravity vector & $\boldsymbol{\phi}_{b}$ \\
Body linear and angular velocities &  $\boldsymbol{v}_b, \boldsymbol{\omega}_b$ \\
Joint positions and velocities & $\boldsymbol{q}, \boldsymbol{\dot{q}}$\\
Previous actions & $\boldsymbol{a}_{t-1}$ \\
Height map scan & $h$ \\
Desired end-effector pose & $\boldsymbol{p}_{ee}^{\ast}$\\
\hline
\end{tabular}
\end{center}
\end{table}

\subsubsection{Reward function for Whole-Body Locomanipulation}\label{sec:reward}

The task-space end-effector pose tracking scenario has two main problems that need to be addressed when formulating the reward function, which is the key part that leads the policy to learn a desired behavior. 

The first problem is that a mobile manipulation task like ours is inherently a two-stage hierarchical approach: We first need to use the locomotion capabilities of our system to get close enough to the target pose, so that this is within reach of our manipulator. Then, we need to maintain a stable base position that allows the manipulator to reach the desired EE pose. In literature, these tasks are usually handled separately, since they are both complex problems by themselves, and combining them yields additional difficulties \cite{fu2022deep, portela2025wholebodyendeffectorposetracking}. However, by separating the locomotion and manipulation tasks, we don't leverage the WBC capabilities of our system that might help in some crucial tasks, such as positioning the quadruped in a way that helps the manipulator reach a difficult pose, or using the manipulator while the base is moving to save valuable operation time or even helping the base to balance in rough terrain.

The second problem is related to the pose tracking task itself. Successful 6D tracking implies precisely reaching a 3D position and 3D orientation. Both goals have usually been framed separately in the reward formulation, with separate position and orientation tracking rewards. Although a valid approach, this introduces the need to balance both terms so that precise tracking happens for both magnitudes. Balancing these terms is not straightforward since the magnitudes of the errors are different, and precisely tracking a 3D orientation has proven to be specially challenging \cite{portela2025wholebodyendeffectorposetracking}. 

\paragraph{Whole-Body 6D EE Tracking Reward} To address both of these challenges, we propose a reward formulation that implicitly handles the locomotion and manipulation phases, as well as inherently balances position and orientation tracking such that no term is prioritized above the other. 

To compute the pose reward, we first compute the position and orientation rewards. The position reward is defined as the sum of the squared norms of the difference between current and desired positions in base frame. The orientation reward is the magnitude of the quaternion difference between current and desired orientations in base frame as well:
\begin{equation}\label{pos-and-rot-rewards}
    e_{pos} = \Sigma||p_{ee} - p_{ee}^{*}||^{2}, e_{rot} = quat_{ee} \ominus quat_{ee}^{*}
\end{equation}

where the difference between quaternions is computed as the magnitude of the axis-angle representation of the relative rotation between quaternions. Then, the individual rewards for position and orientation are computed as:
\begin{equation}
    r_{pos} = e^{\left( -\frac{e_{pos}}{\sigma_{pos}}\right)}, r_{rot} = e^{\left( -\frac{e_{rot}}{\sigma_{rot}}\right)}
\end{equation}

where $\sigma_{pos}, \sigma_{rot}$ are parameters to adjust the sensitivity of the exponential decay with respect to the error. Lastly, to inherently balance these two terms, we compute the pose tracking reward as the product of both terms:
\begin{equation}\label{eq:pose-rew}
    r_{pose} = r_{pos} \cdot r_{rot}
\end{equation}

This way, increasing the reward for tracking one part at the cost of sacrificing the other won't yield higher rewards. Instead, a steady increase in both position and orientation tracking is the only way to maximize the reward function.

\paragraph{Legged Locomanipulation Reward} The reward formulation in \cref{eq:pose-rew} doesn't leverage the mobility of the quadrupedal robot that serves as mobile base of our locomanipulation system. This could cause the policy to learn overly aggressive arm behaviors, as it doesn't have a explicit reward signal informing that moving the base closer to the EE target also helps the manipulator reach such desired pose.

To provide the policy with this information, we formulate a body-to-target gated reward, which encourages the robot body to be close to the desired EE cartesian position up to a radius, $r$, from which the arm can comfortably reach the commanded pose. Considering $d_{base} = ||p_{body, xy} - p_{ee,xy}^{*}||$, then the base position tracking reward is defined as follows:
\begin{equation}
    r_{base} = e^{\left(-\frac{d_{base} - r}{0.5}\right)}
\end{equation}

where the $xy$ suffix indicates that we only use the components in the XY plane to compute the difference. The combined base and EE pose tracking reward is as follows:
\begin{equation}\label{eq:task-reward}
    r_{task} = r_{pose} \cdot (g \cdot r_{base})
\end{equation}
where $g = Sigmoid(k \cdot(d_{base} - r)) \in (0.0, 1.0)$ is the gating factor that modulates the importance of the base distance reward. The intuition is that the locomanipulator first needs to get to a distance $r$ from the target EE position in XY plane, and once it is inside this radius, then it can get the maximum value of the pose reward term. 

With this formulation, we encourage the policy to first make the robot move towards a position near the desired target, and then inherently balance the position and orientation tracking to achieve a successful 6D pose control. Furthermore, formulating these different stages in a continuous manner, without explicit state machines and thresholds, favors learning and leads to a smooth and autonomous transition between task stages.

\subsubsection{Power minimization rewards}

Lastly, we also want to induce the learning agent to gravitate towards the actions that result in the lowest possible energy consumption while still achieving the desired goal. In the RL literature, this is typically handled by the means of \textit{penalty} functions, which are basically negative rewards that discourage the policy of taking certain actions. However, due to the constrained method that we employ here (detailed in \cref{sec:background}), a negative reward would make that violating the constraints would indeed help the agent in minimizing reward loss, since the $(1 - \delta)$ coefficient would be reducing the penalty given to the agent. 

In order to adapt to this characteristic, we flip the penalty formulation so that instead of reducing the reward when power is high, we give higher rewards when the power consumption is low. We do this by applying an exponential kernel to the sum of squared norms of the mechanical power exerted by each joint, such that:
\begin{equation}
\begin{aligned}
    r_{power} = \omega_{legs} \cdot e^{\left(-\frac{\Sigma||\boldsymbol{\dot{q}}_{legs} \cdot \boldsymbol{\tau}_{legs}||}{\mu_{legs}}\right)}\\ + \omega_{arm} \cdot e^{\left(-\frac{\Sigma||\boldsymbol{\dot{q}}_{arm} \cdot \boldsymbol{\tau}_{arm}||}{\mu_{arm}}\right)}    
\end{aligned}
\end{equation}

where $\omega_{legs}, \omega_{arm}$ are the reward weights for each term, $\mathbf{\dot{q}}$ and $\mathbf{\tau}$ represent the joint velocities and torques for both the legs and the arm joints, and $\mu$ is a normalization parameter corresponding to an average maximum power of each joint group.

\subsection{Constraints for Safe and Robust Locomanipulation in Lunar Environments}

With the main task reward already defined, we focus now on setting the constraints that we want our policy to satisfy while learning the task described by the reward function. These constraints can have both real physical meaning (joint limits, maximum torques...) or describe a desired behavior for our robot, such as not exceeding a certain stepping force, surpassing a body orientation threshold, or moving within some velocity limits. 

Following the formulation of previous works \cite{chane2024cat}, we define two different types of stochastic constraints: Hard constraints, where a violation results in immediate termination of the cumulative reward from that point on, and Soft constraints, where violating such constraints results in a probabilistic termination of part of the future reward. The conceptual difference between these constraints is that there are critical situations that we must avoid every time, such as colliding with the environment with a part of the robot that is not a foot, but there are other circumstances that we want to avoid, but we can allow some temporal constraint violation if by all means necessary (e.g. exceeding the established velocity limit).

Taking into account the platform used, the task at hand, and the restrictions and safety guarantees that operation in Lunar environments imposes, we define the following set of constraints employed by our policy during the learning process.

\paragraph{Soft constraints} We define soft constraints $c_{q_{j}}, c_{\dot{q}_{j}}, c_{\tau_{j}}$ to constrain joint positions, velocities, and torques within its operational range and maximum nominal values (i.e. we want to avoid peak torques and velocities, even if physically achievable). In addition, we include style constraints to achieve desired base motion. We set $c_v$ to limit body linear velocity to a maximum of 0.25 m/s, and $c_{rot}$ with a limit of 0.3 radians to avoid excessive base rotation. Finally, $c_{f_{std}}$ constrains the maximum standard deviation in the distribution of forces applied by each foot, thus favoring an even mass distribution across feet. This is crucial in our Lunar operation scenario, where feet might sink in regolith present on the moon surface and provoke catastrophic failure.

\paragraph{Hard constraints} We define a set of hard constraints to specify situations that the policy must avoid while performing the desired task. We set a constraint $c_{contact}$ that restricts high contact forces in any part of the robot that is not its feet. We also set a $c_{fall}$ constraint that is triggered when the body orientation in roll or pitch angles exceeds 90º, to avoid base positions that would most likely cause a fall and damage the robot. Following the same reasoning, a $c_{h, min}$ representing the minimum body height allowed is set to prevent the body from hitting the ground. Lastly, to account for the exceptional situations that Lunar operation poses, we add two additional constraints specifically tailored for operation in Lunar environment. We set $c_{f,max}$ as the maximum impact force that can be exerted by the robot feet, to avoid high-force impacts that might make the robot lose contact with the ground, floating up to dangerous heights, or damage the legs causing an emergency stop. Similarly, we set the $c_{h,max}$ constraint to limit the maximum height that the robot must admit during operation, allowing for some floating and jumping motion leveraging low-gravity conditions on the moon, but not getting too far from the surface to avoid losing control of the situation.

%% file: sections/results.tex
\section{EXPERIMENTAL RESULTS}\label{sec:results}

\subsection{Implementation details}\label{sec:implementation}

We train our locomanipulation policy for reaching a desired end-effector 6D pose while satisfying a user-defined set of constraints. The platform employed for our experiments consists of a Unitree Go2 quadrupedal robot with an Interbotix WX250s 6 DoF manipulator mounted on top of it introduced in \cref{fig:intro}. To emulate the hazardous conditions of lunar exploration, we reduce gravity magnitude to 1/6 th of Earth's gravity, and generate rough, uneven terrain that resembles the morphology of lunar surface.   

At each training episode, we sample a target EE position in task space within a cylinder of 1.2m radius, and 0.7m height centered at the robot's XY position. The desired orientation is sampled uniformly within a $\pm30$ degrees range from the default EE orientation. We include domain randomization techniques to robustify the control policy. Specifically, at each reset we randomize the mass of the quadruped within $\pm10\%$ of its nominal value. We also apply a delayed PD controller to track the desired joint positions provided by the policy, with a random delay of maximum 40ms is applied to the control action. Lastly, we add gaussian noise to the observations received by the policy, in order to account for non-perfect state estimation and sensor measurements. 

Regarding constraint probabilities, for the soft constraints we define a probability curriculum like the one in \cite{dadiotis2025dynamicobjectgoalpushing}, where each constraint has a minimum and maximum termination probability, increasing linearly as the training progresses. We set the maximum value for these constraints at 60\% of the full training time, allowing for a more exploratory behavior at the start of the training run, and becoming more restrictive as training advances and base skills have been learned. The minimum and maximum probabilities for each constraint term can be seen in \cref{tab:constraint_satisfaction}. 

We use NVIDIA Isaac Sim \cite{makoviychuk2021isaacgymhighperformance} as the high-fidelity physics simulator for our task, and NVIDIA Isaac Lab \cite{mittal2023orbit} as the learning framework to create our learning environment. We implemented a customized version of the constrained PPO algorithm in \cite{chane2024cat} by modifying the PPO implementation in RSL-RL library \cite{rudin2022learning}. The policy $\pi_{\theta}$ is parametrized by a MLP with three hidden layers of 512, 256, and 128 neurons each, with ELU activation functions\cite{clevert2016fastaccuratedeepnetwork}. Each training episode lasts 10s if not terminated earlier, with the policy running at 100Hz and the underlying PD controllers running at 200Hz. We train our policy for 10000 learning iterations with 4096 parallel environments with one robot each, which takes approximately 5h in a desktop machine with a single NVIDIA GeForce RTX3090 GPU.

\subsection{Results} 

\begin{table*}[ht!]
\caption{Constraint Probabilities and Experimental Constraint Satisfaction results}
\label{tab:constraint_satisfaction}
\begin{center}
\begin{tabular}{c||c|c|c|c|c|c|c|c|c|c|c}
\hline
\textbf{Constraint} & $\mathbf{c}_{q_j}$ & $\mathbf{c}_{\dot{q}_j}$ & $\mathbf{c}_{\tau_j}$  & $\mathbf{c}_v$  & $\mathbf{c}_{rot}$ & $\mathbf{c}_{f_{std}}$ & $\mathbf{c}_{contact}$ & $\mathbf{c}_{fall}$ & $\mathbf{c}_{h,min}$ & $\mathbf{c}_{h,max}$ & $\mathbf{c}_{f,max}$ \\
\hline \hline
\textbf{Min. Prob.} & 0.05 & 0.05 & 0.05 & 0.05 & 0.05 & 0.05 & 1.0 & 1.0 & 1.0 & 1.0 & 1.0 \\
\hline
\textbf{Max. Prob.} & 0.9 & 0.9 & 0.25 & 0.25 & 0.9 & 0.25 & 1.0 & 1.0 & 1.0 & 1.0 & 1.0 \\
\hline
\textbf{Episode Violation Time \%} & \textbf{0.002} & \textbf{0.0} & \textbf{0.004} & \textbf{0.08} & \textbf{0.02} & \textbf{0.005} & \textbf{0.0} & \textbf{0.0} & \textbf{0.0} & \textbf{0.0} & \textbf{0.0}
\\
\hline
\end{tabular}
\end{center}
\end{table*}

\begin{figure}[t]
    \centering
    \begin{subfigure}{0.45\textwidth}
        \includegraphics[width=\textwidth]{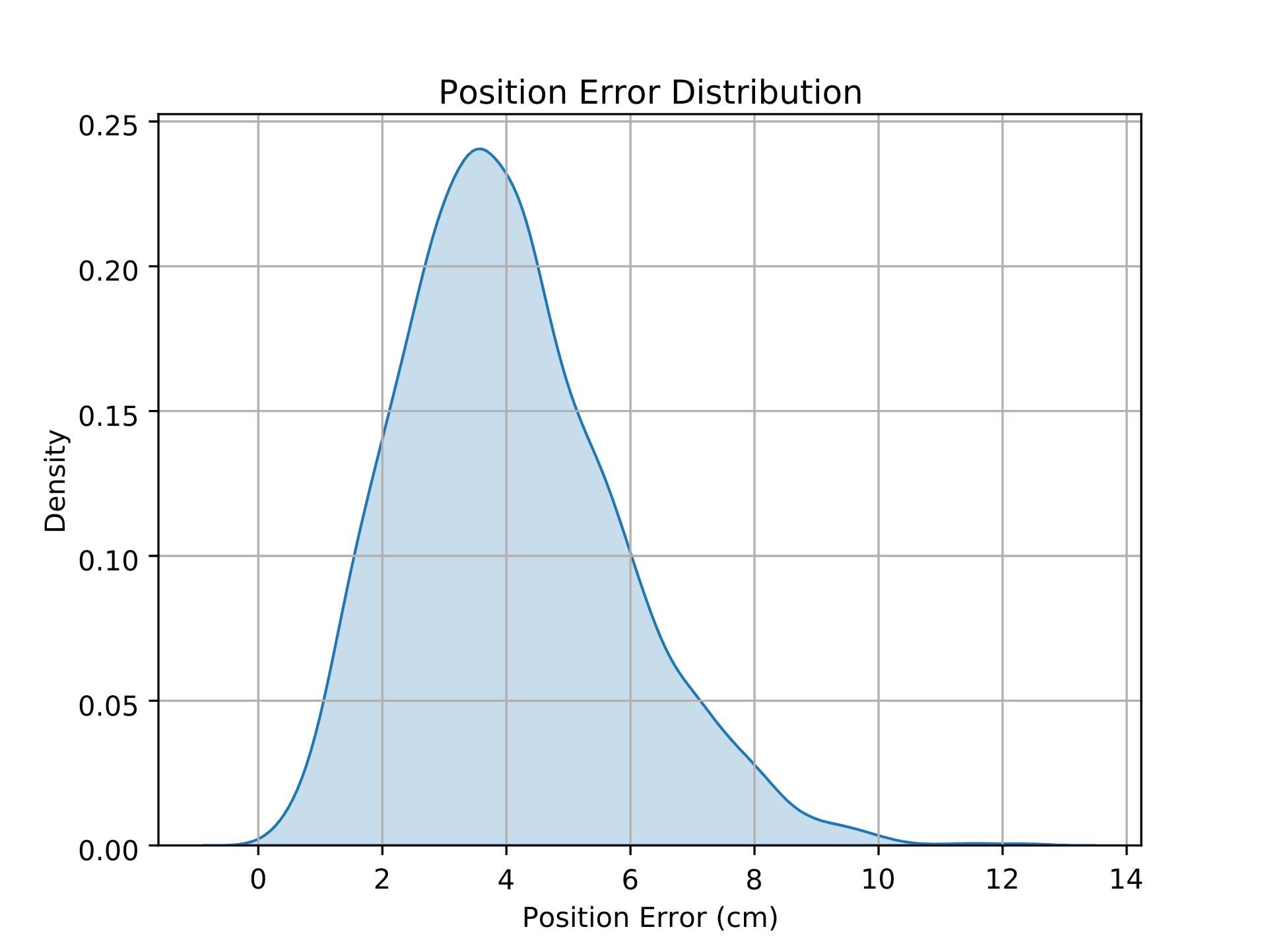}
        \label{fig:subfig1}
    \end{subfigure}
    \begin{subfigure}{0.45\textwidth}
        \includegraphics[width=\textwidth]{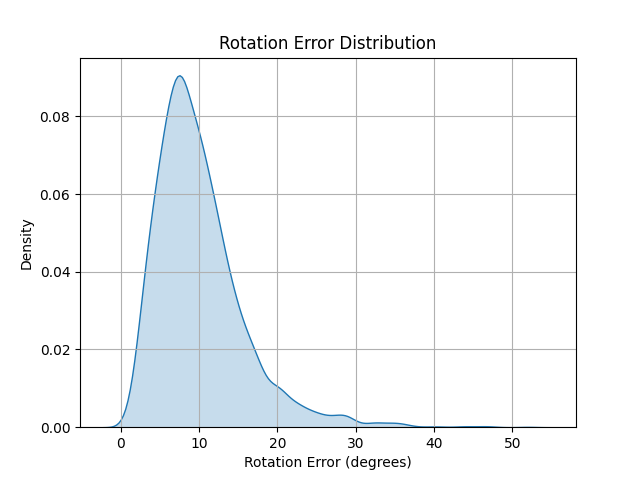}
        \label{fig:subfig2}
    \end{subfigure}
    \caption{Distribution of the position and orientation errors for the 4096 evaluated samples}
    \label{fig:error_dist}
\end{figure}

\begin{figure*}[ht]
    \centering
    \includegraphics[width=0.95\linewidth]{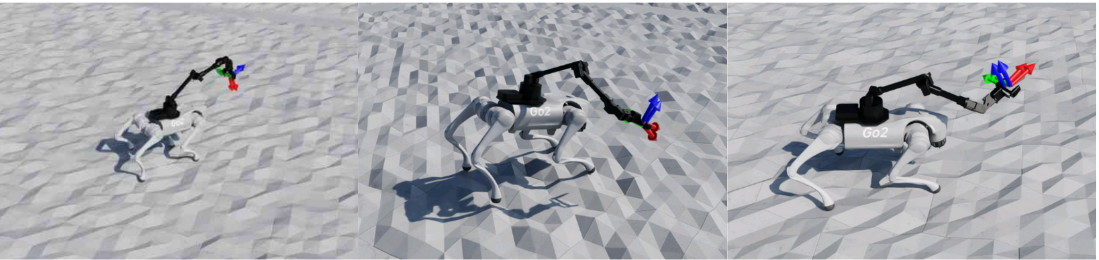}
    \caption{\textbf{Whole-Body EE Pose Tracking Examples.} Our legged mobile manipulator shows precise and safe EE pose tracking in difference configurations, allowing its use for tasks such as sample collection or surface inspection.}
    \label{fig:pose-tracking}
\end{figure*}

We perform our simulation experiments by running a single episode across 4096 environments, and report the mean final pose tracking error to assess the performance of our whole-body EE control, and the average proportion of time that each constraint is not satisfied across all testing environments to evaluate the effectiveness of the employed CRL formulation.

\paragraph{End-effector task-space pose tracking} We sample the target poses for evaluating pose tracking performance as described in \cref{sec:implementation}, and compute the error in the same manner that was described for the reward functions in \cref{sec:reward}. We report an average position error of 4cm, and orientation error of 8.1 degrees for our locomanipulation task, a performance similar to state-of-the-art locomanipulation works in terrestrial domain \cite{portela2025wholebodyendeffectorposetracking, ha2024umilegs}. The position and orientation error distributions are shown in \cref{fig:error_dist}, where it can be seen how the maximum position errors are around 10cm, while the majority of the density is concentrated between 2-6cm position errors. In \cref{fig:pose-tracking}, a series of poses tracked by the robot are shown, illustrating how our legged locomanipulator successfully employs all its DoF in order to reach the desired pose. In addition to this, emergent behaviors that take advantage of low-gravity conditions are shown by our policy, where slight jumping gaits appear in order to minimize the power consumption of the quadruped joints, while using small movements of the arms and legs to stabilize during these maneuvers.

\paragraph{Constraint satisfaction} We measure constraint satisfaction as the average proportion of episode time that each constraint is violated, as in previous works \cite{dadiotis2025dynamicobjectgoalpushing, chanesane2024solo}. \cref{tab:constraint_satisfaction} shows the average percentage of episode that each constraint is not satisfied. We can see how all hard constraints are always respected by the policy, ensuring robust and safe deployment with respect to our most restrictive decisions. In terms of soft constraints, successful behavior is also observed as the mean percentage of time that each constrained is violated is $\leq 0.08\%$ in the worst case, with an average percentage of $\sim0.01\%$ for all soft constraints.

%% file: sections/conclusions.tex
\section{CONCLUSIONS}

In this paper, we introduced a constrained reinforcement learning (CRL) framework specifically designed for quadrupedal mobile manipulators operating in challenging lunar environments. The proposed method successfully integrates whole-body locomotion and manipulation, leveraging explicit safety constraints including collision avoidance, dynamic stability, and power management.

Experimental evaluations in simulation demonstrated the method's outstanding performance in tracking 6D end-effector poses with an average positional error of just 4 cm and an orientation error of approximately 8.1 degrees, matching the performance standards established by state-of-the-art terrestrial systems. Additionally, the CRL framework showed exceptional constraint satisfaction, maintaining near-perfect compliance with both hard and soft constraints across extensive testing.

Notably, our system displayed emergent behaviors that exploited the low-gravity lunar conditions, adopting efficient and adaptive motions to minimize energy consumption while preserving task effectiveness. These results underscore the potential of the proposed CRL-based approach for enabling reliable, adaptive, and safe autonomous mobile manipulation in lunar exploration missions, effectively addressing the critical operational challenges inherent to space robotics.

Future work should seek to evaluate the policy in more challenging terrains, such as craters or steep hills. In addition, real-world experiments in lunar-analog terrains in combination with task planning algorithms would be relevant to validate the adaptability of our work and showcase the potential of constrained reinforcement learning in challenging real-life scenarios.